\documentclass[11pt]{article}

\usepackage[final]{acl}

\usepackage{times}
\usepackage{latexsym}

\usepackage[T1]{fontenc}

\usepackage[utf8]{inputenc}

\usepackage{microtype}

\usepackage{inconsolata}

\usepackage{graphicx}

\usepackage{booktabs}
\usepackage{enumitem}
\usepackage{amsmath}
\usepackage{amssymb}
\usepackage{xcolor}
\usepackage[ampersand]{easylist}
\newcommand{\gt}{y}            
\newcommand{\pred}{\hat{y}}    

\title{Star Light, Star Bright: Can Multimodal Models Parse an Astronomy Visual Question Answering Dataset Right?}

\author{J. P. Naiman \\
  School of Information Sciences \\
  University of Illinois, Urbana-Champaign, IL 61820, USA \\
  \texttt{jnaiman@illinois.edu}}

\begin{document}
\maketitle


\begin{abstract}

    Figures are essential for scientific communication, and the field of Visual Question Answering (VQA) aims to test the figure understanding capabilities of large multimodal models (LMMs).  Here, we investigate the accuracy of several low-cost LMMs on domain-specific VQA tasks for astronomical \textit{images of the sky}. 
    As instrument for this task we build a domain-specific VQA dataset comprising 2001 partially-synthetic figures divided evenly between contour plots, real telescope images drawn from NASA's \textsf{SkyView}, and ``fake'' sky images which share the axes and angular scales of the real images but draw their colors from a gaussian mixture model, alongside 47357 questions. 
    Each figure is released as pristine and with realistic page-aging degradations mimicking the historical astronomical corpus. Testing three low-cost proprietary LMMs, we find models frequently fail to answer domain-specific questions at high accuracy. 
An initial comparison with 104 human-verified, astrometrically tagged figures from historical  journal publications shows similar median errors in angular extent. All code is available on GitHub: \url{https://github.com/jnaiman/SkyImagesWASP2026} and dataset is available on HuggingFace: \url{https://huggingface.co/datasets/ReadingTimeMachine/astro_sky_image_vqa}.

\end{abstract}

\section{Introduction}

In academic articles, figures synthesize and interpret scientific information. Their centrality has made figure parsing a main task of pipelines that digitize scientific documents \cite[e.g., document layout analysis,][]{binRev2019,ding2026deep,kumar2025exploring}, alongside figure classification \cite{jobin_docfigure_2019} and dataset extraction \cite{davila_chartinfographics_2019}. 
With the growing prevalence of large multimodal models (LMMs) in scientific practice, much recent work tests whether these models can parse figures and ``understand'' their meaning \cite{hoque_chart_qa_survey_2022,farahani2023} -- commonly through Visual Question Answering (VQA), in which a model is given an image and questions about the image. 
Initially focused on natural images \cite[e.g.,][]{vqafornatural}, chart-based VQA is now an expanding field \cite[e.g.,][]{huang_pixels_2024}. Early studies covered simple bar, line and scatter charts \cite{kahou2018figureqaannotatedfiguredataset}, while follow-up work added more complex chart types \cite[e.g., funnel, ][]{han2023chartllama} and plots whose relationship to the underlying data is indirect, such as histograms and contour plots \cite{naiman_what_2026,tpdl2026}. 
Absent are domain-specific VQA tasks, which are necessary to probe realistic scientific LMM applications \cite{chang2022mapqa,huang_pixels_2024}.

Astronomy relies on \textit{images of the sky}, a figure type that would benefit from domain-specific VQA. Sky images share some complexity with contour plots, mapping light intensity in a wavelength range to a color rather than a spatial scale \cite{tpdl2026}, but encode position in right-ascension (RA) and declination (DEC) -- a non-Cartesian system that can deviate substantially from a rectangular grid over large sky areas. 
Historical sky figures (published before $\sim$1997) add a further complication: many are grayscale rather than color, and scanning for upload to digital archives introduces errors of its own \cite{grant_ads_data_holdings_2000,baird_degradation_2007}.

To the author's knowledge, there are no VQA datasets centered on images of the sky, for either present-day or historical figures. However, benchmarks focused on astronomy have recently been developed for other AI and machine learning (ML) tasks.  
\textsf{AstroVisBench} \cite{joseph_astrovisbench_2025} evaluates whether models can write the code that produces astronomical visualizations, and \textsf{AstroMMBench} \cite{shi_astrommbench_2025} evaluates domain-specific reasoning over curated astronomical imagery via multiple-choice questions. However, neither measures LMM free-form answers requiring calculations with a known ground truth.

Our research question is: \textit{can low-cost LMMs perform domain-specific parsing of astronomical sky figures -- including distinguishing real and synthetic sky sources -- and does this degrade for historical-style figures?} To answer it, we present a VQA dataset focused on images of the sky in both present-day and historical contexts, alongside two visually similar plot types: contour plots drawn from several two-dimensional distributions, and ``fake'' sky images that keep their RA/DEC axes but draw their colors from a gaussian mixture model (GMM) rather than from a telescope. 
The data is used to test low-cost proprietary LMMs, comparing accuracy between original (``pristine'') figures and those modified by realistic aging processes. Code, images, question sets and LMM answers are released for use by the community.

\section{Dataset Construction}

\subsection{Figure Generation}

Single-panel figures are generated following the work of \citet{naiman_what_2026,tpdl2026} and are rendered with \textsf{matplotlib} \cite{hunter_matplotlib_2007}. 
Style, colormap, DPI, aspect ratio, font and font size are randomly selected; titles and axis labels are drawn from a word pool mined from the ar$\chi$iv, with inline mathematics substituted at a fixed rate \cite{naiman2023Ocr}. 
Alongside each image the underlying data array, sampled generation parameters, and pixel-coordinate bounding boxes for the plot area, axes, tick labels, title, axis labels, colorbar and colorbar label are saved. 
Figures are rejected and re-sampled if labels run off the canvas, if bounding boxes overlap beyond a tolerance, or if the plot area falls below a fraction of the figure. 

Figure types are sampled evenly between contour plots with colors from various distributions (random, linear, gaussian mixture model) \cite{tpdl2026}, images of the sky taken from telescopes, and images of the sky created by a gaussian mixture model.   
The GMM-sky images serve as a control: real-sky images do not provide figures with realistic axes, angular scale, rendering and color-scale distribution without a  telescope signal. Comparing answers between the two families isolates whether errors stem from the domain-specific RA/DEC axis conventions (shared by both) or from the image source (which differs). 

For both types of images of the sky, axes are stored as world coordinate system (WCS) with RA/DEC tick labels in several styles and a sampled epoch (e.g., J2000, B1950). 
A (target, wavelength) pair draws from a set of objects from the astronomical literature (targets described in  \citealt{naiman_targets_inprep}), and retrieved from sky surveys within NASA's \textsf{SkyView} \cite{mcglynn_skyview_1996} through \textsf{astroquery} \cite{ginsburg_astroquery_2019}, with WCS handling via \textsf{astropy} \cite{astropy_2022}.

The physical size of an image centered on a target on the sky can change across surveys and applications. 
To mimic a realistic scale of a sky image shown in an article, when SIMBAD \cite{simbad} sky-size for images (angular field of view, FOV) is available for a target, we use this to set the range of RA/DEC from which to extract an image. When absent, we make use of the FOV and angular resolution information stored in \textsf{SkyView} for each sky survey and select RA/DEC ranges from 50 times the smallest resolution up to 0.75 times the FOV. GMM-generated sky images follow the same distribution of angular sizes as the real sky images. 
Real-sky cutouts are cached at $300\times300$ and square. To create a set of realistic aspect ratios (and to match with contour plot parameters), the grid is sampled ($n_x$ from 50--300, $n_y = n_x/\mathrm{aspect}$). The index map between initial and rescaled pixels is exactly affine, such that arcsec/pixel stays isotropic and the center is preserved. 
A logarithmic stretch with clipping at the 99th percentile and its inversion is applied to both GMM- and real-sky images -- the net effect removes extreme outliers while leaving the bulk of the distribution on its original scale. 

Units are not included in colorbars. Only $\sim$20\% of retrieved real-sky images declare a form of physical units (e.g., \textsf{BUNIT} -- Jy/beam, MJy/sr, K, Rayleighs), and typically several data processing steps are applied between raw sky image data and published real-sky figures \cite[e.g.,][]{starck_astronomical_2006}.
A fuller pipeline of image processing \cite[e.g.,][]{joseph_astrovisbench_2025} to more closely align with the domain science (and thus, physical colorbar units) is left as future work. 

Our synthetically generated figures trade the visual realism of a human-collected and annotated figure dataset for the detail of access to underlying data and the precision of bounding box locations   \cite{methani_plotqa_2020,wang_charxiv_2024,masry_chartqapro_2025,naiman_what_2026}.   
Our inclusion of telescope-derived images of the sky with realistic angular scales acts as a compromise between a fully-synthetic and a hand-annotated figure dataset.

\subsubsection{Page Aging of Figures} \label{sec:aging}

The figure generation pipeline described previously produces pristine scientific charts; however, much of the astronomical literature is not ``born digital'', but instead consists of human-scanned copies of originally paper-only journal articles. 
Given that the storage and scanning procedures for historical images can degrade the uploaded images, we also build a copy of each figure with several realistic ``page aging'' image augmentations.  
We make use of the \textsf{Augraphy} \cite{groleau_augraphy_2023} package, which models damage as accumulating in three phases -- \textit{ink} (e.g., ink bleed), \textit{paper} (e.g., color, texture) and \textit{post} (e.g., stains, folding). 
Our application pulls randomly from this set of effects following \citet{guan_prepocr_2025} to mimic a modified form of the ``archive'' set of effects from \citet{scansimulator_2025}, which is meant to align with expected degradations resulting from scanning historical documents. As many of the historical articles in the field of astronomy were scanned at high resolution by trained archivists \cite{grant_ads_data_holdings_2000}, the uploaded documents retain a lower level of page augmentations -- denoted as ``\textsf{archive-light}'' in this paper.  
Finally, we transform 2/3 of the images to grayscale to mimic the color palette of historical articles. 
Not all of the randomly selected original colormaps are suitable to transform to grayscale which adds a further source of error in the aged figures.

\begin{figure}[h]
  \centering
  \includegraphics[width=0.9\columnwidth]{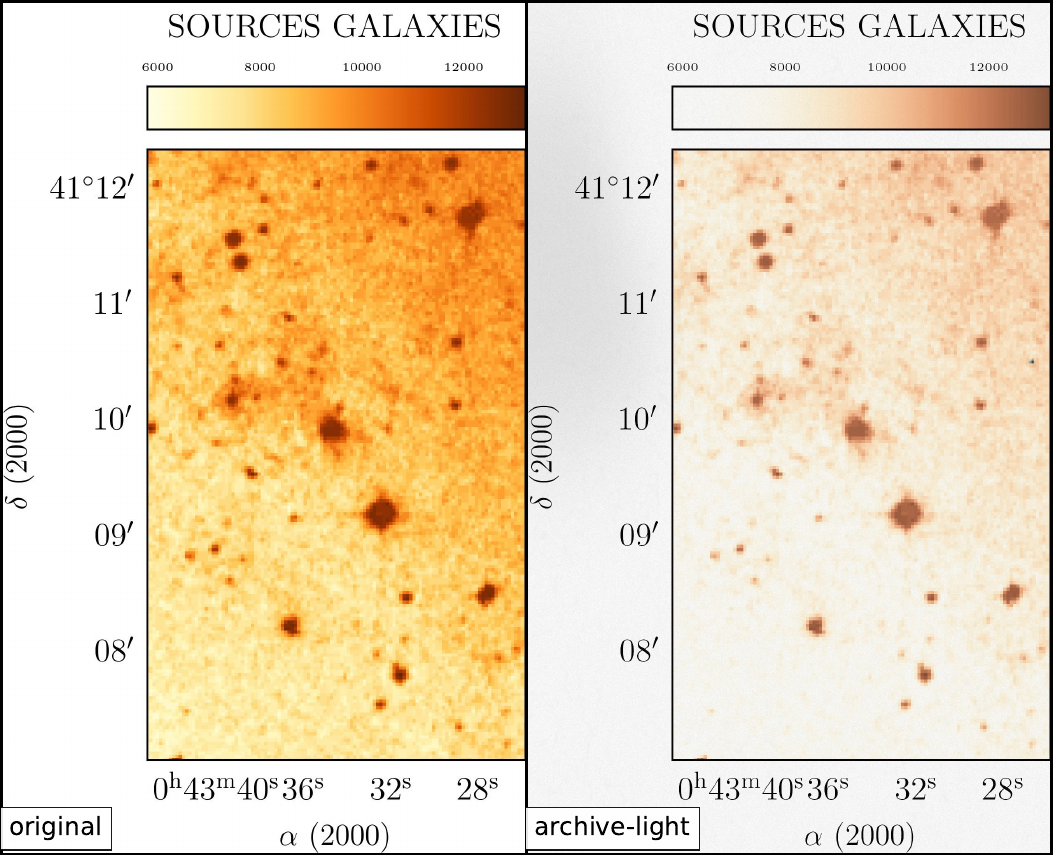}
  \caption{A dataset figure before (left) and after (right) \textsf{archive-light} aging. 
  }
  \label{fig:aging}
\end{figure}

An example of a real-sky image (left panel) and aged counterpart (right panel) is shown in \autoref{fig:aging}.

\subsection{Question Generation}

Questions are generated for each figure using the four-level complexity model of \citet{lundgard_accessible_2022}, as adapted to chart-based VQA in other works \cite{tang_vistext_2023, naiman_what_2026, tpdl2026}. We generate questions at Levels 1--3, spanning what is directly readable off the figure, what requires reading values, and what requires inference beyond a single statistic. Level 4 requires external context or domain expertise and is out of scope for this work. Example questions are listed in \autoref{tab:questions}; \autoref{app:questions} gives further detail and the accessibility motivation for this question scaffolding.

\section{Results and Analysis}

From our initial dataset of 2001 images, evenly spread between contours, real-sky images, and GMM-sky images and a total of 47357 questions, we randomly select 255 images (85 in each category) for evaluation with three proprietary LMMs, yielding 6035 questions per model. 
We use each in a low-cost configuration: \textsf{GPT-5.4-nano} \cite{openai_gpt54_2026} at default reasoning and verbosity (\textsf{GPT}), \textsf{Gemini-3.5-flash-lite} \cite{google_gemini_31_pro_2026} with \textsf{thinking\_level = minimal} (\textsf{Gemini}), and \textsf{Claude-Haiku-4.5} \cite{anthropic_claude_haiku_45_2025} with thinking disabled (\textsf{Claude}). We restrict this preliminary run to the low-cost tier, where \citet{naiman_what_2026} found accuracy varied most.  These are each provider's lower-cost reasoning setting and are not strictly equivalent across providers.  
The models selection and question scale assumes a cost-sensitive archiving institution, while holding the 95\% Wilson interval \citep{kahouadji2025comprehensive} on a single accuracy to at most $\pm$10.4 percentage points (pp) with 85 figures per category, and $\pm$7.4 pp for the 170 pooled sky images -- enough to resolve the large differences discussed below, but not fine-grained model rankings. 
Questions are issued one per API call, so each answer is an isolated zero-shot observation. Each model is run over two conditions --- the original figures and \textsf{archive-light}. 
Answers are requested as JSON with entries for raw answer and explanation.  
Only 48 of the 36210 answers (0.13\%) fall outside this pair and are unparsable at scale -- 21 that are not valid JSON and 27 explicit \texttt{null} refusals; per-model counts are given in \autoref{app:parsing}. 
Here, we report question types that probe sky-image features 
-- plot type, provenance and angular scale in RA/DEC. 

\begin{figure}[t]
  \includegraphics[width=\columnwidth]{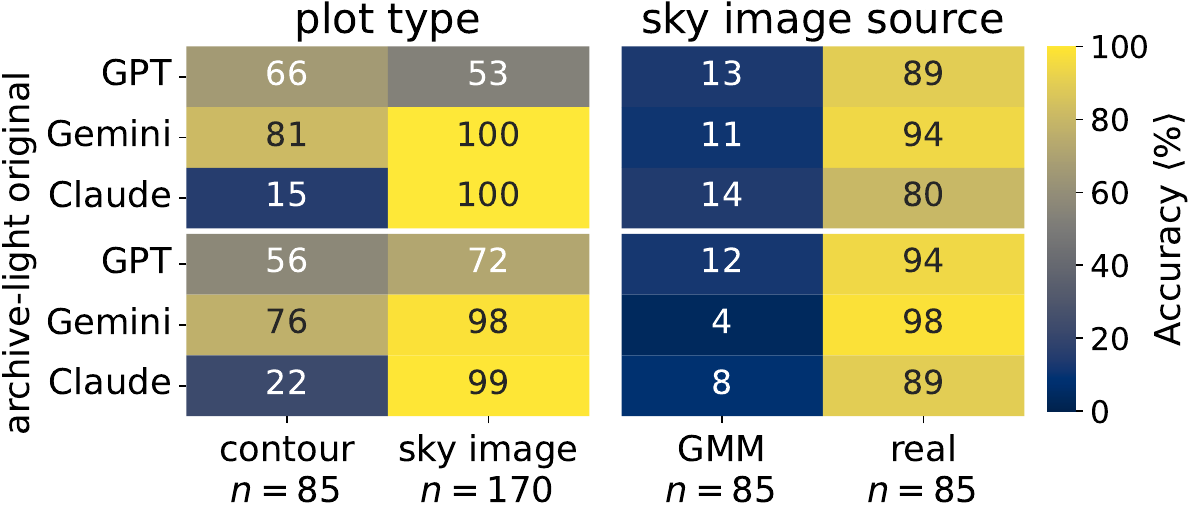}
  \caption{Accuracy of identifying plot type (left) and real vs generated sky (right) for the three low-cost models. Models are offered explicit choices for each. } 
  \label{fig:plottype}
\end{figure}

\paragraph{Plot type and Provenance.} \autoref{fig:plottype} shows the accuracy of answers to  ``What is the plot type of the plot in this figure? Please choose the plot type from the following list: [contour, image of the sky].'' and, for sky images, ``What is the underlying distribution used to create the data in this figure? Please choose the distribution from the following list: [gaussian mixture model, real image of the sky].'' 

These questions test two sources of possible confusion in the models -- the presence of an image-based figure (contour) and the presence of the specific x/y labels of RA/DEC (GMM-sky). 
The tests are motivated by work showing scholarly documents can be a carrier for problematic adversarial content aimed at AI-assisted pipelines \cite{lin_hidden_2026,nagaraja_image_2026}.  
GMM-sky images are sampled to the same pixel, angular and color-scale distributions as the real ones, with parameters chosen to span the range from point sources to diffuse emission. We did not pursue statistical indistinguishability 
as the aim was to remove obvious artifacts while preserving the structural differences the task is meant to probe. 
A gradient-boosted classifier on low-level image features (e.g., field size) separates real from GMM-sky images with 85\% accuracy (\autoref{app:classifier}). The two families are thus distinguishable from image content, so low LMM accuracy reflects model capability rather than an impossible task. 

The accuracy shown in \autoref{fig:plottype} is measured as a simple binary (yes/no) for both original and \textsf{archive-light} historical figures. 
The left, plot type panels of \autoref{fig:plottype} show that when confronted with a contour plot \textsf{Claude} does worse than guessing (15\% original, 22\% \textsf{archive-light}) and \textsf{GPT} only modestly better (66\% and 56\%), except \textsf{Gemini} which shows $\approx$80\% accuracy for both original and \textsf{archive-light} figures. \textsf{Gemini} and \textsf{Claude} recover sky-image plot types almost always, but this is to be expected from the over-guessing of this type of figure evident in the contour plot panels; \textsf{GPT}, by contrast, labels only 53\% of original sky images correctly. For sky image source, the models clearly favor guessing ``real image of the sky'' and frequently miss when the image source is a GMM distribution.
Only \textsf{Claude} returned unparsable answers here (\autoref{app:parsing}). 
Across models and conditions, GMM-sky images are identified correctly only 4--14\% of the time, against 80--98\% for real-sky images. 
With $n=85$ per cell, differences of $\lesssim$10 pp lie within the 95\% interval of a single cell, but the largest gaps do not: for example, \textsf{Claude}'s 15\% versus \textsf{Gemini}'s 81\% on contour plots gives $z=8.6$ ($p<10^{-16}$, simple two-proportion test).

Our dataset includes open-ended (no lists provided) versions of several questions. For example, in the plot type question raw accuracy is near zero for all three models as they answer with strings such as ``astronomical survey image'' that describe the figure correctly and never match the ground-truth label. Further parsing of open-ended answers is left for future work.

\begin{figure}[t]
  \includegraphics[width=0.98\columnwidth]{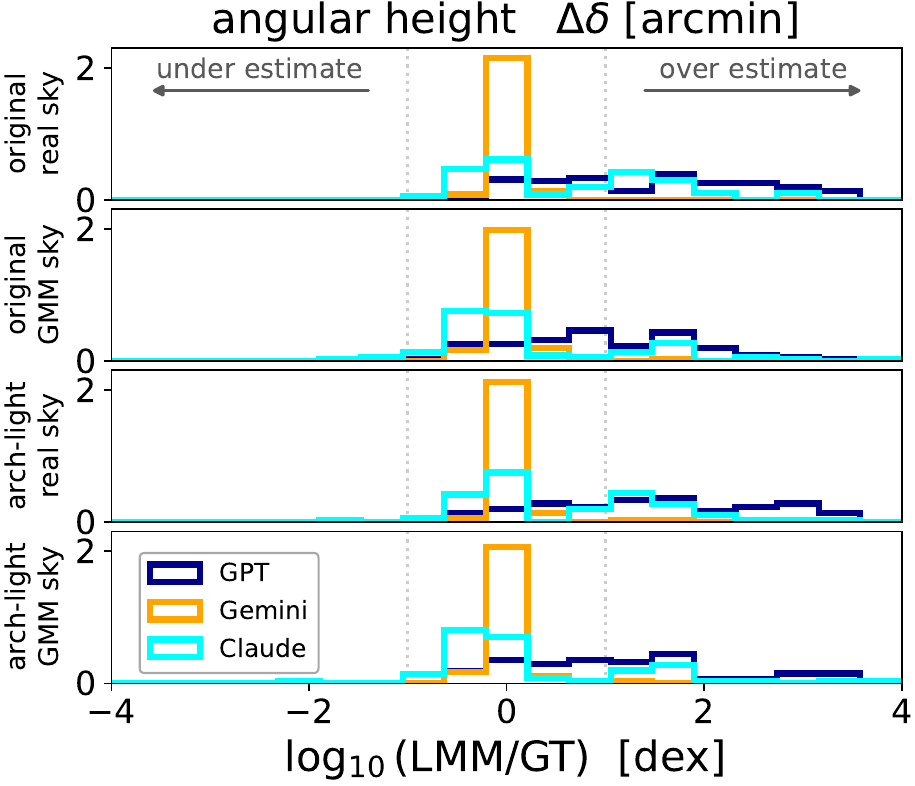}
  \caption{Error distribution of angular height for GMM- and real-sky images measured as log-ratio, $\mathcal{R}$. Zero indicates agreement with ground truth, $\pm$1 indicates predictions too large or too small by a factor of ten. }
  \label{fig:stats}
\end{figure}

\paragraph{Angular height.} \autoref{fig:stats} shows the accuracy of answers to the question ``What is the angular height of the sky region shown in this figure, measured along the declination axis? Give the value in arcminutes.'' This requires reading the y-axis of the figure and translating the domain-specific measure of angular size (see \autoref{fig:aging}) into another typical unit of angle in astronomy (arcmin, 1/60$^\circ$).

The accuracy is measured by the log-ratio, $\mathcal{R} = \log_{10}(\pred/\gt)$, defined for $\gt > 0$ and $\pred > 0$ and scales the magnitude of a prediction similarly for under or over estimates. 
The metric cannot be used when LMM and ground truth (GT) answers have different signs which occurs in our dataset for a single GPT, \textsf{archive-light} answer.   

Questions about angular scale probe how much detail is expected (angular resolution) and what other targets fall within the field (angular height/width). 
While \textsf{Gemini} does the best overall (peaked around $\mathcal{R}\approx 0$ in \autoref{fig:stats}), 
the other models have larger spreads tending toward overestimation with \textsf{GPT} having 53\% (60\%) of its responses at $>+1$~dex and \textsf{Claude} having 32\% (32\%) for pristine (arch-light/\textsf{archive-light}) figures. Initial parsing of explanations returned by the models suggests the slight overestimations (0.5 < $\mathcal{R}$ < 1) stem from imprecise calculation of the y-axis ranges and rounding of answers. Larger errors ($\mathcal{R}$ > 1) often result from misreading of the y-axis tick marks due to domain-specific norms (e.g., the bottom y-tick in \autoref{fig:aging} is read as $08^\circ 00'$ instead of $08'$). 
Splitting \autoref{fig:stats} by source shows the GMM-sky images are potentially easier for this axis-reading question: on the original figures, \textsf{Claude} overestimates by more than a factor of ten significantly more often on real-sky than on GMM-sky figures (41\% vs.\ 24\%, $p = 0.01$), and \textsf{GPT} shows the same direction, though not significantly (59\% vs.\ 47\%, $p = 0.12$).


\paragraph{Published figures.} To test whether results on our figures carry over to published ones, we collect sky images with human-verified astrometry from the AAS Astronomy Image Explorer (AIE) \cite{aie_2014,adap}, which indexes figures from the Astrophysical and Astronomical Journals and includes Astronomy Visualization Metadata (AVM) in the WorldWide Telescope convention \cite{rosenfield_wwt_2018}. The AVM stores central RA/DEC and pixel scale, but does not include figure bounding box. Following an initial pass in OpenCV for potential image corners \cite{bradski_opencv_2000,suzuki_border_1985,douglas_peucker_1973} bounding boxes and figure canvas spans were verified by hand using Label Studio \cite{tkachenko_labelstudio_2020}.  Along with automated checks of AVM to SIMBAD's position of the object named in its publication's title (if available) and checks for coordinate duplication or nonphysical field sizes, human verification  reduced the AVM-tagged set of figures from 163 to 104.

\begin{figure}[t]
  \includegraphics[width=\columnwidth]{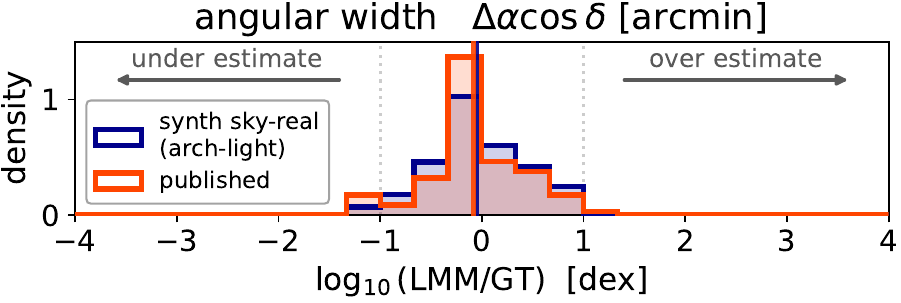}
  \caption{Log-ratio error, $\mathcal{R}$, of \textsf{Gemini}'s angular-width answers for real-sky figures (\textsf{archive-light}, blue; $n=85$) and for 103 
  human-verified published figures from the Astronomy Image Explorer (orange). } 
  \label{fig:realsky}
\end{figure}

As AVM tags exclude information for other figure elements (e.g., colorbar), questions are limited to angular scale and passed to our best-performing (\textsf{Gemini}) model. \autoref{fig:realsky} compares angular-width errors (change in RA de-projected from spherical coordinates, see \autoref{app:angular}). Answers (103, 1 unparsable) are similar across synthetic (real-sky) and published figures: median $\mathcal{R}=-0.08$~dex against $-0.05$~dex for \textsf{archive-light}, with 55\% vs.\ 51\% of answers within a factor of two.

\section{Conclusions and Future Work}
\label{sec:conclusions}

We have introduced a domain-specific chart VQA dataset for astronomical figure types: contour plots and images of the sky. 
Sky panels are drawn both from real telescope observations and from GMMs, generated for equal numbers in contour, real- and GMM-sky image categories. We include both pristine images and ``aged'' images to mimic those in the historical astrophysics corpus.

Across three proprietary low-cost LMMs we find figure-type identification often fails to distinguish between sky and contour plots and, for sky images, between real and GMM images. 
Models tend to overestimate the angular heights of sky regions. 

On a small set of human-verified published figures, \textsf{Gemini}'s angular-width answers match its median behavior on our partially synthetic figures.  
We detect no difference in accuracy between pristine and aged images apart from one cell (\textsf{GPT} labels aged sky images correctly more often -- 53\% vs.\ 72\%, $p<0.001$).  
At $n=85$ per model/category we can exclude only aging effects larger than $\sim$21 pp at 80\% power. Similar questions asked with less domain-specific charts do not display such large inaccuracies \citep[e.g.,][]{naiman_what_2026}.  These results motivate the need for the development of further domain-specific VQA datasets \cite{chang2022mapqa,huang_pixels_2024,naiman_what_2026}. %

Future work could focus on a more domain-accurate image processing pipeline, target other types of astronomy-specific figures (e.g., spectra), and investigate the explanations and open-ended answers stored as part of the LMM responses in our dataset which we did not explore here.

\clearpage

\section*{Limitations}

The figures are generated, not collected. The arrays inside the real sky panels are genuine survey cutouts, but the figure drawn around them is not a published figure, and lacks conventions real ones carry --- domain-standard colormaps, compass roses, beam ellipses, overlaid catalog positions. 
Our published-figure comparison suggests angular extent transfers in the median, but it covers only one model and two questions (angular width, and angular height in \autoref{app:angular}). %
The 104 AVM-tagged figures are also a small, non-random sample of the sky images in the historical literature -- those an archive happened to tag with astrometry \cite{adap} -- so the comparison should be read as an example of how results on our figures carry over, not as a definitive measure of LMM accuracy on published sky figures.

Matching the synthetic sky fields to the real ones removes the obvious cues to provenance, but does not make the two families indistinguishable and was not intended to. The provenance question we pose is therefore hard rather than impossible, and the accuracies we report should be read against a task that remains somewhat solvable in principle from image content alone (see \autoref{app:classifier}).

We evaluate only the lower-cost configuration of three proprietary models, on a single aging preset and these configurations are not strictly equivalent in reasoning effort across providers. Open-weight models, higher-reasoning configurations and astronomy-tuned models are untested, and model versions and prices change quickly enough that these particular numbers have a short shelf life.

Each condition is a single run over 255 of the 2001 figures. Per-cell uncertainties are up to $\pm$10.4 pp, so small differences between models, or between pristine and aged figures, should not be over-interpreted. 

Our published-figure comparison is small (104 human-verified figures, one model) and limited to the geometric questions that astrometric metadata can ground; colormap, provenance and value questions have no ground truth for figures we did not draw. Its angular extents rely on human-verified plot boxes, and pixel scale cannot be compared at all, because the published astrometry counts scan pixels rather than data pixels. We do not include a human baseline on the real-versus-GMM question, which would require a hand-annotated corpus beyond the scope of this short paper. 

The dataset is designed as an evaluation benchmark. We do not test the use of the synthetic images for training or data augmentation, though it is not outside the realm of possibility that this dataset could be used as training or pre-training for a variety of models.

Scoring is exact match after normalization for categorical answers and a log-ratio for numeric ones. As the open-ended plot-type results show in our preliminary investigation, exact match understates a model that is substantively right in unanticipated vocabulary. The questions themselves are generated from templates over known parameters: correct by construction, but not always a sample of what a working astronomer would ask.

\section*{Ethics Statement}

The sky cutouts used here are public data retrieved from NASA's SkyView service and are redistributed within the dataset in derived form, with full provenance metadata.
The published figures used for comparison may remain under their journals' copyright; we release only their metadata, questions and model answers, with a script that retrieves the images from the Astronomy Image Explorer. The synthetic sky panels in this dataset are visually plausible but are not observations, and are labeled as such in every record; we note this explicitly because the same matching procedure that makes them useful as a control would make unlabeled synthetic sky images misleading if separated from their metadata.

More broadly, image falsification is a recognized form of scientific misconduct\footnote{\url{https://bps.stanford.edu/home/instances-scientific-misconduct/falsification-images}} (a growing issue along with adversarial image and text content injection \cite{lin_hidden_2026,nagaraja_image_2026}), and our synthetic sky images are, by design, plausible enough that the models we test usually label them as real. The procedure that makes them a useful control could be misused to fabricate observations in a manuscript, and our results suggest that low-cost LMMs 
would not reliably flag them. We mitigate this by labeling every synthetic image in its record and releasing the generation code and parameters so that any image can be traced, and we encourage use of the dataset to develop provenance detectors rather than as a source of figures, though we cannot limit the codebase's use toward generation of synthetic figures passed off as authentic.

AI was used to port code from prior work to the current repository and help to provide documentation. AI was also used to write analysis and verification code (e.g., the classifier baseline, statistical checks, and table and cost generation, partially adapted from author-written code), all of which was reviewed by the author. It was additionally used to assist in the literature review (all sources verified by humans), with a draft of the methods section, appendices, and the abstract, and with the writing of the Limitations and Ethics Statement section. It was also used to help shorten the paper to fit page requirements and to help draft revisions responding to reviewer comments; all revised text was checked by the author. AI was trialed for other parts of the writing, but was not found useful in other sections so was abandoned for the main section of the paper. 

\bibliography{custom}

\appendix

\section{Answer Parsing}
\label{app:parsing}

Answers are requested as JSON with entries for the raw answer and an explanation. A small number fall outside that pair and cannot be parsed at scale. Out of 6035 questions per model, this occurred for 12 (22), 4 (3) and 5 (2) answers for \textsf{GPT}, \textsf{Gemini} and \textsf{Claude} respectively, where the first number is for the original images and the bracketed number for \textsf{archive-light}. For \textsf{Gemini}'s and \textsf{Claude}'s are responses that are not valid JSON: most break on the quote characters of sexagesimal tick labels (e.g., \verb|12'00"|), and four of \textsf{Claude}'s answer the provenance question in free text with no JSON at all. 
27 of \textsf{GPT}'s 34 are valid JSON with an explicit \texttt{null} answer, typically for color statistics it judges unrecoverable from the image. For the plot-type and provenance questions of \autoref{fig:plottype} specifically, only \textsf{Claude} returned unparsable answers: 3 for the original images and 1 for \textsf{archive-light}, out of 425 each.

\section{Question Levels}
\label{app:questions}

Questions follow the four-level figure description model of \citet{lundgard_accessible_2022}, adapted to chart-based VQA in \citet{tang_vistext_2023,naiman_what_2026,tpdl2026}. Level 1 covers what is directly readable off the figure (e.g., plot type); Level 2 requires reading values off the axes or colorbar (e.g., the median of the data along an axis, or the angular height of the field); Level 3 requires inference not reducible to a single statistic, such as which distribution family the underlying data came from or multi-step, complex, calculations. Level 4 requires external context or domain expertise and is not generated here.

Beyond quantifying question complexity, this scaffold matters for accessibility: many scientific figures remain unusable for blind and low-vision readers \cite{wang_improving_2021}, and Levels 2 and 3 are the tiers those readers report as most useful \cite{lundgard_accessible_2022, seo_maidr_2024, kim_exploring_2023}.

\autoref{tab:questions} gives example questions for our figures; \autoref{tab:published_questions} lists every question asked of the published figures and where its ground truth comes from.

\begin{table*}[t]
\centering\small 
\begin{tabular}{@{}c p{0.56\textwidth} p{0.16\textwidth} p{0.17\textwidth}@{}}
\toprule
Lvl & Question [choices offered] & Applies to & Example Answer \\
\midrule
1 & What is the plot type of the plot in this figure? [contour, image of the sky] & all plots & image of the sky \\
1 & What is the colormap that was used in this figure? & all plots & rainbow\_r \\
1 & What coordinate epoch is given on the axis labels of this figure? Answer ``none'' if no epoch is stated. & GMM-sky, real-sky & J2000 \\
1 & What is the smallest unit of angle shown on the right ascension tick labels of this figure? [hours, minutes, seconds] & GMM-sky, real-sky & minutes \\
1 & What is the minimum right ascension value covered by the right ascension axis of this figure -- that is, the lower limit of the axis range, not of the data? Give the value in hours, minutes and seconds [...] & GMM-sky, real-sky & 18h46m50.98s \\
1 & What is the style of the contour plot? [image, contour lines, both] & contour & contour lines \\
\midrule
2 & What is the angular height of the sky region shown in this figure, measured along the declination axis? Give the value in arcminutes. & GMM-sky, real-sky, published & 1.58 \\
2 & What is the angular width of the sky region shown in this figure, measured along the right ascension axis, as a true angle on the sky (i.e.\ including the cos(declination) factor)? Give the value in arcminutes. & GMM-sky, real-sky, published & 1.88 \\
2 & What is the mean value of the data along the color-axis in this figure? & all plots & 108.4 \\
\midrule
3 & What is the underlying distribution used to create the data in this figure? [gaussian mixture model, real image of the sky] & GMM-sky, real-sky & real image of the sky \\
3 & What is the underlying distribution used to create the data in this figure along the color-axis? [random, linear, gaussian mixture model] & contour & linear \\
3 & Approximately what angular size does a single pixel of this sky image span? Give the value in arcseconds. & GMM-sky, real-sky & 188.0 \\ 
3 & What is the right ascension of the centre of this sky image, in the J2000 (FK5) frame? [...] Give the value in decimal degrees. & published & 91.95 \\
\bottomrule
\end{tabular}
\caption{Example questions for each level (Lvl column), the plot types they are asked of, and an example ground-truth answer. Choices offered are given in brackets; long prompts are abridged ([...]). Plot types: contour, GMM-sky and real-sky are our figures; published are the human-verified Astronomy Image Explorer figures. Example answers come from single figures of each type (\textsf{vqa\_000001}, GMM-sky; \textsf{vqa\_000002} and \textsf{vqa\_000004}, real-sky; \textsf{vqa\_000007}, contour; \textsf{rsky\_000001}, published); numeric answers are in the units the question asks for.}
\label{tab:questions}
\end{table*}

\begin{table*}[t]
\centering\small 
\begin{tabular}{@{}c p{0.42\textwidth} p{0.24\textwidth} p{0.13\textwidth} c@{}}
\toprule
Lvl & Question & Ground truth & Range & Shared \\
\midrule
2 & What is the angular width of the sky region shown in this figure, measured along the right ascension axis, as a true angle on the sky (i.e.\ including the cos(declination) factor)? Give the value in arcminutes. & mean length of the plot box's top and bottom edges on the sky & 0.33--957$'$ & yes \\
2 & What is the angular height of the sky region shown in this figure, measured along the declination axis? Give the value in arcminutes. & mean length of the plot box's left and right edges on the sky & 0.33--1429$'$ & yes \\
3 & What is the right ascension of the centre of this sky image, in the J2000 (FK5) frame? [...] Give the value in decimal degrees. & sky position of the plot-box centre & 23--346$^\circ$ & no \\
3 & What is the declination of the centre of this sky image, in the J2000 (FK5) frame? [...] Give the value in decimal degrees. & sky position of the plot-box centre & $-$67--75$^\circ$ & no \\
\midrule
3 & Approximately what angular size does a single pixel of this sky image span? Give the value in arcseconds. & not released: scan pixels, not data pixels & -- & yes \\
\bottomrule
\end{tabular}
\caption{Questions asked of the 104 human-verified published figures (\textsf{Gemini} only). Ground truth is measured through each figure's AVM astrometry over its human-verified plot box (\autoref{app:angular}); the range is over all 104 figures, in arcminutes ($'$) or degrees. ``Shared'' marks questions worded identically to those asked of our figures, so their scores are directly comparable. Questions about other figure elements (colormap, colorbar statistics, provenance) have no ground truth in the metadata and are not asked; pixel scale was asked but is not scored.}
\label{tab:published_questions}
\end{table*}

\section{Angular Extent of a Sky Figure}
\label{app:angular}

Positions on the sky are given in equatorial coordinates: right ascension $\alpha$, measured eastward along the celestial equator, and declination $\delta$, the angle north or south of it. By convention $\alpha$ is written in hours, minutes and seconds of time, with $24^{\rm h} = 360^\circ$, so that $1^{\rm h} = 15^\circ$, $1^{\rm m} = 15'$ and $1^{\rm s} = 15''$, while $\delta$ is written in degrees, arcminutes and arcseconds. There are two potential issues when reading a sky figure. First, a minute of right ascension is not an arcminute: an axis spanning $2^{\rm m}$ of $\alpha$ covers $30'$. Second, lines of constant $\alpha$ converge toward the celestial poles, so the true angle subtended by an interval $\Delta\alpha$ shrinks with declination; along a line of constant $\delta$ it is $\Delta\alpha\cos\delta$. At $\delta = 60^\circ$, one hour of right ascension spans $7.5^\circ$ on the sky rather than $15^\circ$. Angular height, the extent along the declination axis, needs neither correction, since $\Delta\delta$ is already an angle.

Our ground truth applies the $\cos\delta$ factor at the mid-declination of the displayed field, $w = (\alpha_{\max}-\alpha_{\min})\cos\bar\delta$, whereas for the published figures width is the great-circle length of the plot box's top and bottom edges, measured through the AVM astrometry. The two definitions agree to within 2\% for 94 of the 104 published figures (median 0.4\%) and differ appreciably only for fields spanning several degrees, by up to 23\% for a $16^\circ$-wide field at $\delta \approx 58^\circ$.

\begin{figure}[t]
  \includegraphics[width=\columnwidth]{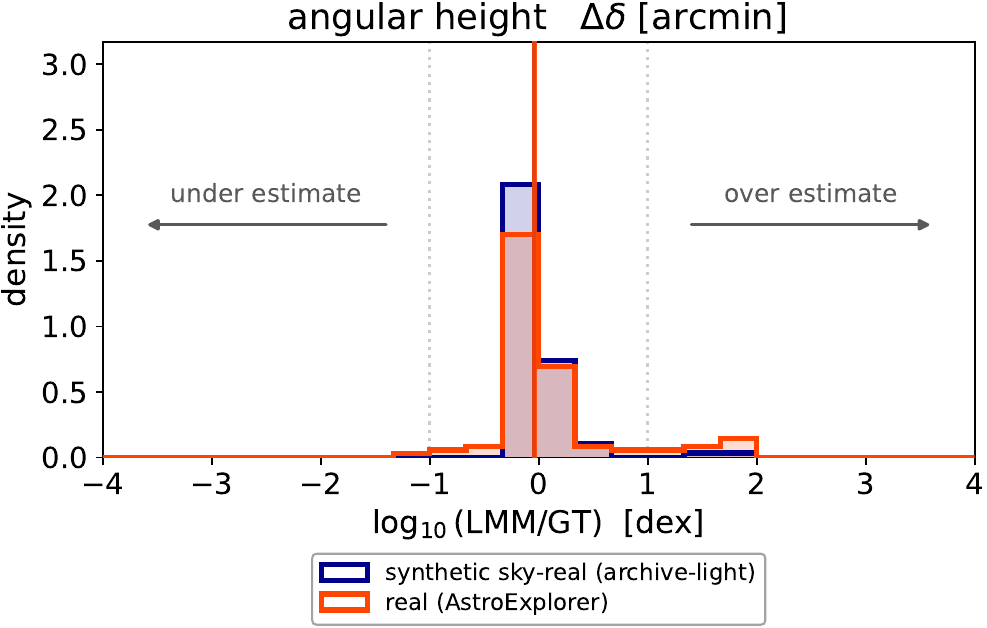}
  \caption{As \autoref{fig:realsky}, for angular height: log-ratio error, $\mathcal{R}$, of \textsf{Gemini}'s answers for our real-sky figures (\textsf{archive-light}, blue; $n=85$) and for the 104 human-verified published figures (orange).}
  \label{fig:realsky_height}
\end{figure}

\paragraph{Angular height on published figures.} \textsf{Gemini} was also asked the angular height of each published figure, worded identically to our figures (\autoref{tab:published_questions}). \autoref{fig:realsky_height} compares the errors. In the median the two sets agree, as for width: $\mathcal{R} = -0.04$~dex for published figures against $-0.05$~dex for \textsf{archive-light}. Published figures scatter more, however: 76\% of answers fall within a factor of two, against 94\% for our figures ($p < 0.001$, two-proportion test), and 10\% are too large by more than a factor of ten, against 2\%. Since height needs no $\cos\delta$ correction, this extra scatter cannot come from the right ascension conventions above.

\section{Classifier Baseline for Provenance}
\label{app:classifier}

To determine if the provenance question can be answered from image content, we train a gradient-boosted classifier \cite{scikit-learn} to separate real-sky from GMM-sky figures using three low-level features of each figure's data array: the field size (the geometric mean of its angular width and height), the concentration of its emission (the share of the background-subtracted signal held by the brightest 5\% of pixels), and the sign of the data minimum. Over all 1334 sky figures (667 of each family), 5-fold cross-validation gives 85\% accuracy, against 50\% for always guessing one family. Most of this comes from concentration, which alone reaches 80\%: point-source-dominated real fields are more concentrated than the GMM fields. Field size and the sign of the minimum carry little information on their own, as the matching of the two families intends.

We leave out a fourth feature, the color span (the dynamic range of the data). GMM fields borrow their color ranges from real cutouts, and in this release the same ranges recur across many GMM figures: 667 GMM figures share 56 distinct spans, while real figures have 473. A classifier can exploit that regularity across the dataset -- color span alone reaches 81\% -- but no single figure reveals it, so it says nothing about whether the provenance question is answerable from one image. The code is released with the dataset generation code with updates to pull a color range distribution more closely resembling that of the real-sky images intended in future releases.

\section{Archive Aging Details}
\label{app:aging}

\autoref{tab:preset} lists the \textsf{archive-light} preset  implemented: 11 effects, their \textsf{Augraphy} phase, the independent probability with which each fires for a given figure, and the realized firing rate over the 255-figure corpus. Effects marked \emph{soft} run with parameters pulled back from the \textsf{Augraphy} defaults --- ink bleed at intensity $0.1$--$0.3$, ink mottling at alpha $0.05$--$0.12$, paper hue restricted to $28$--$40$ at low saturation, stain blending at alpha $0.18$, shadows at opacity $0.05$--$0.2$, rotations restricted to $\pm1^\circ$ rather than $\pm3^\circ$, and JPEG quality held at $70$--$95$. \textsf{Folding} and \textsf{SubtleNoise} run at their defaults but fire less often.

Relative to the full \textsf{archive} preset of \citet{scansimulator_2025}, three effects are dropped entirely --- \textsf{BleedThrough}, \textsf{LightingGradient} and \textsf{PageBorder}. A further four effects available in \textsf{Augraphy} are excluded from both \textsf{archive} \citep{scansimulator_2025} and our \textsf{archive-light} presets: \textsf{Scribbles} and \textsf{WaterMark} add marks unrelated to the figure, \textsf{BookBinding} bends the page until the plot is no longer readable as a plot, and \textsf{Letterpress} restyles the ink against the colormap.  None of these are likely to apply to our figures as they do not typically mimic damage expected in archive scans \citep{grant_ads_data_holdings_2000}.

\begin{table}[h]
\centering\small
\begin{tabular}{llcc}
\toprule
Phase & Effect & $p$ & fired \\
\midrule
ink   & InkBleed \emph{soft}            & 0.50 & 55\% \\
ink   & InkMottling \emph{soft}         & 0.30 & 34\% \\
paper & ColorPaper \emph{soft}          & 0.80 & 80\% \\
paper & NoiseTexturize \emph{soft}      & 0.50 & 55\% \\
paper & BrightnessTexturize \emph{soft} & 0.50 & 49\% \\
post  & Stains \emph{soft}              & 0.40 & 38\% \\
post  & Folding                         & 0.25 & 25\% \\
post  & ShadowCast \emph{soft}          & 0.25 & 24\% \\
post  & Geometric \emph{soft}           & 0.50 & 49\% \\
post  & SubtleNoise                     & 0.50 & 48\% \\
post  & Jpeg \emph{soft}                & 0.60 & 59\% \\
\bottomrule
\end{tabular}
\caption{The \textsf{archive-light} preset: firing probability and the rate realized over 255 figures.  Each aged figure is also converted to grayscale with probability 2/3 (realized: 65\%).}
\label{tab:preset}
\end{table}

For each figure we record the sampled plan --- the effects that fired, their phase assignment, the grayscale flag, the page background color and the seed --- in a manifest released with the dataset, so any aged image can be regenerated without redistributing it. The plan is recorded rather than recomputed at read time, so a manifest stays valid for the images it describes even if the preset is later changed.

\section{Cost Analysis}
\label{app:cost}

\autoref{tab:cost} gives the cost of running each model once over the evaluation, i.e.\ every question asked a single time per condition. For \textsf{Gemini} and \textsf{Claude} it is computed exactly from the token counts recorded in each API response; one question costs about \$0.0007 with \textsf{Gemini} and \$0.0025 with \textsf{Claude}. Our code does not recorde \textsf{GPT}'s token usage, and its reasoning tokens are billed but not returned, so its cost is scaled from the provider's billed cost for its final runs: \$3.34 for the 4970 calls covering the last 105 figures of in each condition, or about \$0.0007 per question.  All costs use non-batch API pricing so larger datasets would likely be less expensive. 

\begin{table}[h]
\centering\small
\begin{tabular}{@{}llrrr@{}}
\toprule
Model & Condition & In (M) & Out (M) & Cost (\$) \\
\midrule
GPT & both (est.) & -- & -- & 8.11 \\
Gemini & original & 7.54 & 0.86 & 4.42 \\
Gemini & archive-light & 7.54 & 0.77 & 4.18 \\
Claude & original & 10.34 & 0.96 & 15.14 \\
Claude & archive-light & 10.35 & 0.95 & 15.11 \\
\midrule
Gemini & published & 0.66 & 0.20 & 0.69 \\
\bottomrule
\end{tabular}
\caption{Cost of one run of each model: tokens in and out (millions) and cost in USD at list prices per million input / output tokens (GPT \$0.20 / \$1.25; Gemini \$0.30 / \$2.50; Claude \$1.00 / \$5.00), checked 12 October 2026. Each condition is 6035 calls per model; the published-figure run is 520 calls (\textsf{Gemini} only). GPT, whose responses do not record token usage, is scaled from its provider's billed cost for the last 4970 calls (both conditions).}
\label{tab:cost}
\end{table}

\end{document}